# FarExStance: Explainable Stance Detection for Farsi

Majid Zarharan[◇,•], Maryam Hashemi[†,•], Malika Behroozrazegh[•]
Sauleh Eetemadi[∗], Mohammad Taher Pilehvar[♠], Jennifer Foster[◇]
[◇] School of Computing, Dublin City University
[†] Iran University of Science and Technology
[•] Dadmatech, Tehran, Iran
[∗] School of Computer Science, University of Birmingham
[♠] School of Computer Science and Informatics, Cardiff University
majid.zarharan2@mail.dcu.ie

## Abstract

We introduce FarExStance, a new dataset for explainable stance detection in Farsi. Each instance in this dataset contains a claim, the stance of an article or social media post towards that claim, and an extractive explanation which provides evidence for the stance label. We compare the performance of a fine-tuned multilingual RoBERTa model to several large language models in zero-shot, few-shot, and parameter-efficient fine-tuned settings on our new dataset. On stance detection, the most accurate models are the fine-tuned RoBERTa model, the LLM Aya-23-8B which has been fine-tuned using parameter-efficient fine-tuning, and few-shot Claude-3.5-Sonnet. Regarding the quality of the explanations, our automatic evaluation metrics indicate that few-shot GPT-4o generates the most coherent explanations, while our human evaluation reveals that the best Overall Explanation Score (OES) belongs to few-shot Claude-3.5-Sonnet. The fine-tuned Aya-32-8B model produced explanations most closely aligned with the reference explanations.

## 1 Introduction

The task of stance detection refers to the process of determining the position or stance of a piece of text towards a claim or target. For example, given the perspective *Another athlete has tragically lost their life to COVID-19* and the claim *No athlete has died from COVID-19 to date*, the stance towards the claim is *Disagree*. Stance detection is a useful step towards automated claim verification – an increasingly pressing challenge, given the exposure of individuals to misinformation online. *Explainable* stance detection is a form of the task where an explanation is supplied along with the stance label. Most stance detection research to date has been conducted on English.

We introduce FarExStance, the first Farsi dataset designed specifically for stance detection tasks, which includes extractive explanations as evidence. Our aim in building this dataset is to address the gap in resources for stance detection and explainable NLP in Farsi. FarExStance comprises 5,874 unique claims generated by annotators based on headlines and news summaries collected from over 100 Farsi news agency websites. These claims were then used to gather 26,307 instances from three different sources: Farsi news agencies, Twitter (now X), and Instagram. Each instance was manually classified into one of four categories: *Agree*, *Disagree*, *Discuss*, or *Unrelated*, following the standards set by Pomerleau and Rao. (2017) and Zarharan et al. (2019). Additionally, annotators provided sentence-level evidence to support their classifications. The resulting dataset is a versatile resource that can be used for a range of tasks, including (explainable) stance detection, evidence retrieval, fact-checking, and summarization.

To illustrate the challenges presented by FarExStance, we build an explainable stance detection baseline using XLM-RoBERTa-Large (Conneau et al., 2019). We also experiment with Large Language Models (LLMs), making use of zero- and few-shot prompting, parameter-efficient fine-tuning and Retrieval Augmented Generation (RAG) (Lewis et al., 2020). We explore potential biases and provide an estimate of human performance. The highest stance detection accuracy achieved by an LLM is 79.8 compared to a human performance estimate of 79.2. Explanations are more difficult to evaluate, and the LLMs tested vary in their ability to capture important information.

Our novel contributions are: 1) a Farsi stance detection dataset, namely FarExStance, manually curated with labeled instances and accompanying evidence which we make publicly available[1], and 2) benchmark results obtained using multilingual language models of various sizes in zero-shot, few-shot and fine-tuned settings, evaluated through both human and automated evaluation.

---

[1] https://github.com/Zarharan/FarExStance  or
https://github.com/Dadmatech/FarExStance

## 2 Related Work

**Datasets.** Stance detection and fact-checking have seen significant research, with the majority of existing datasets being in English.[2] Prominent English datasets like FEVER (Thorne et al., 2018) and WT-WT (Conforti et al., 2020) are large-scale but lack explanations, limiting their use for explainability. Other datasets provide explanations, whether human-generated in PUBHEALTH (Kotonya and Toni, 2020a), LIAR-PLUS (Alhindi et al., 2018) and EX-FEVER (Ma et al., 2024) or LLM-generated in e-FEVER (Stammbach and Ash, 2020), but still focus exclusively on English-language data. There are a few datasets for Farsi including Persian Stance Classification (Zarharan et al., 2019), ParsFEVER (Zarharan et al., 2021), and Persian Tweets Stance Detection (Bokaei et al., 2022), which focus on news articles, Wikipedia pages, and tweets, respectively. These datasets, while useful, are smaller in scale (ranging from 2.1k to 23k instances) than FARExSTANCE (26.3k instances) and do not provide explanations.

**Stance Detection Approaches.** Several approaches to stance classification have been developed, ranging from feature-based (Qazvinian et al., 2011; Lukasik et al., 2015; Ferreira and Vlachos, 2016; Zeng et al., 2016; Aker et al., 2017; Zhang et al., 2018; Ghanem et al., 2018; Lukasik et al., 2019; Li et al., 2019) to neural network approaches (Kochkina et al., 2017; Chen et al., 2017; Veyseh et al., 2017; Bhatt et al., 2018; Hanselowski et al., 2018; Poddar et al., 2018; Umer et al., 2020).

The advent of pre-trained language models like BERT (Devlin et al., 2019) has significantly influenced advancements in stance detection. For example, Chen et al. (2019) proposed the standard method that involves concatenating the claim and perspective into a single string and feeding it into BERT. Popat et al. (2019) then proposed STANCY, which used cosine similarity between the BERT representations of the claim/perspective pair and the claim in the loss function such that their representations become similar when the perspective supports the claim and dissimilar when it opposes the claim. Yang and Urbani (2021) introduced a BERT-based model called Tribrid, which injected automatically generated negated perspectives into a model to encourage the model to produce more consistent predictions. We use *XLM-RoBERTa-Large* (Conneau et al., 2019) as a baseline in our experiments.

The recent explosion of interest in large language models (LLMs) has prompted research into their use for stance detection across different domains (Ma et al., 2024; Weinzierl and Harabagiu, 2024; Zhao and Caragea, 2024; Zhao et al., 2024; Alghaslan and Almutairy, 2024). One notable approach is the use of Retrieval-Augmented Generation (RAG) models, which are designed to address hallucination in knowledge-intensive tasks by incorporating external knowledge sources (Izacard et al., 2024; Gao et al., 2024; Guan et al., 2024). RAG models have been particularly successful in fact verification, enhancing task accuracy through improved evidence selection (Yue et al., 2024a,b). In this work, we aim to improve LLMs' reasoning capabilities for stance detection by employing the RAG method for more effective evidence retrieval.

## 3 Dataset Collection

We automatically collected political, economic, and sports news published over a six-month period from more than 100 Farsi news agency websites[3]. Inspired by the ParsFEVER dataset (Zarharan et al., 2021), the annotation process consists of two stages: **claim generation** and **claim labeling**. First, we detail the process of generating claims from the collected news headlines and summaries, followed by an explanation of how to determine the stance of various news sources regarding the generated claims (see Figure 1).

### 3.1 Claim Generation

This phase involves generating claims based on headlines and summaries of news articles. Annotators are provided with a random headline and its corresponding summary from the collected news and tasked with generating one to three original claims by paraphrasing the headline, the summary, or both. Since the news is automatically retrieved, some may not be newsworthy, and the resulting claims may not yield relevant results when searched. In such cases, annotators can disregard the item. Additionally, annotators are instructed to mutate the claims. These mutations are altered versions that may either support or contradict the original claim, the headline, or the summary. As in ParsFEVER (Zarharan et al., 2021), we define six types of mutations: paraphrasing, negation, substitution of an entity with a similar or dissimilar one, and making a claim more general or more specific. Furthermore, the latter category is subdivided into two parts: (1) adding one or more words or phrases to the original claim based on annotator knowledge, and (2) using information from the summary. Annotators are required to generate one to three mutated claims for each mutation type (see Appendix B.1 for details).

---
[2]See Table 4 (Appendix) for a comparison of datasets.

[3]See https://github.com/Zarharan/FarExStance for the list of news agency websites.

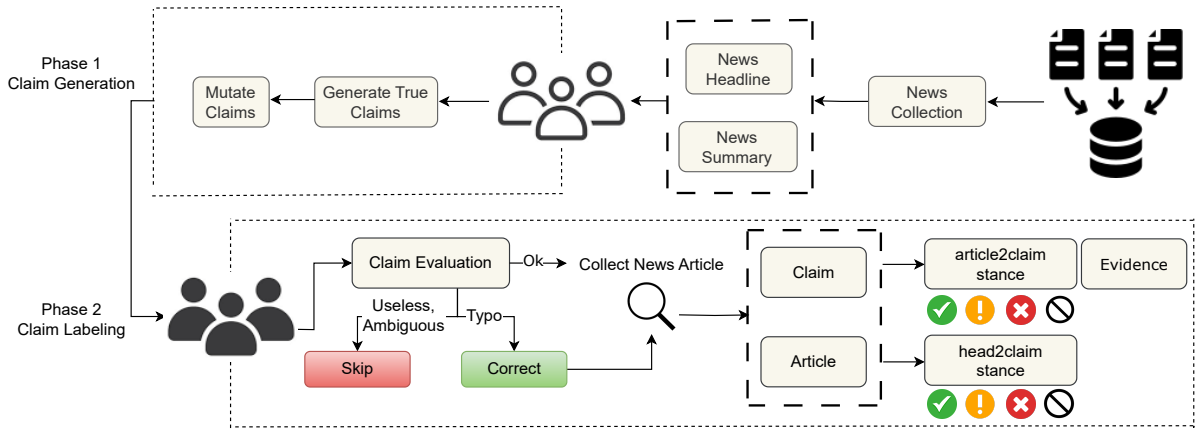

Figure 1: Overview of the Manual Annotation Process. News articles are collected, and true claims are manually generated based on their headlines and summaries. These true claims are then manually mutated, and annotators gather related news articles using search engines for each generated claim. The annotators then label the stance of each instance and provide supporting evidence (Figure 6 in Appendix shows our annotation tool.).

| Stance | English | Farsi |
|---|---|---|
| Disagree | **Claim:** The second vaccine imported by Iran's private sector is AstraZeneca, made in Russia.<br><br>**Perspective:** The spokesperson of the Food and Drug Organization announced the arrival of the first shipment of coronavaccine by the private sector in the country. According to IRNA, Kianoosh Jahanpour wrote on his Twitter account on Thursday: The first shipment of imported vaccine from the private sector of the pharmaceutical sector, including more than 300,000 doses of AstraZeneca vaccine produced in Russia, arrived in the country a few hours ago. | ادعا:<br>دومین واکسن وارد شده توسط بخش خصوصی ایران، آسترازنکا، ساخت روسیه، می باشد.<br><br>متن:<br>سخنگوی سازمان غذا و دارو از ورود اولین محموله واکسن کرونا توسط بخش خصوصی به کشور خبر داد. به گزارش ایرنا، کیانوش جهانپور روز پنجشنبه در حساب کاربری خود در توییتر نوشت: اولین محموله واکسن وارداتی بخش خصوصی حوزه دارو شامل بیش از سیصد هزار دز واکسن آسترزنکا تولیدی روسیه ساعاتی پیش وارد کشور شد. |
| Agree | **Claim:** Protest gathering of Haft Tappeh sugarcane company workers.<br><br>**Perspective:** On Sunday, July 27, the workers of Haft Tappeh Sugarcane company held a protest rally for the sixth day in a row on the premises of this company.<br>In addition to raising their former demands, these workers chanted slogans describing the water shortage situation in Khuzestan as a sign of solidarity with the water uprising in Khuzestan. | ادعا:<br>تجمع اعتراضی کارگران شرکت نیشکر هفت تپه<br><br>متن:<br>روز یکشنبه ۲۷ تیرماه #کارگران شرکت نیشکر هفت‌تپه برای ششمین روز متوالی در محوطه این شرکت تجمع اعتراضی برگزار کردند. این کارگران غیر از مطرح کردن خواسته‌های سابق خود، به نشانه همبستگی با قیام آب خوزستان در تجمع خود شعارهایی در توصیف وضعیت بی‌آبی این استان سر دادند. #اعتراضات |
| Discuss | **Claim:** Crowding of vaccination centers due to the delay in vaccine injection.<br><br>**Perspective:** From the line of chicken to the line of the coronavirus/ Worrying about the lack of vaccine made the elderly line up!<br>With the increase in the vaccination process in the country, reports and the publication of videos and pictures in cyberspace indicate the crowding of the population and the gathering of the elderly and their families, and the formation of long queues in the vaccination centers. | ادعا:<br>شلوغی مراکز واکسیناسیون به دلیل تاخیر در تزریق واکسن<br><br>متن:<br>از صف مرغ تا صف واکسن کرونا/ نگرانی از کمبود واکسن سالمندان را به صف کرد!<br>با افزایش روند واکسیناسیون در کشور گزارش ها و انتشار فیلم و تصاویر در فضای مجازی حکایت از ازدحام جمعیت و تجمع سالمندان و خانواده های آنها و تشکیل صف های طولانی در مراکز واکسیناسیون دارد. |

Figure 2: FarExStance examples, the human-labeled evidence is highlighted in green.

## 3.2 Claim Labeling

In this step, each original claim and its corresponding mutations are assigned to an individual annotator, who is responsible for collecting related news articles and labeling them for stance. Before collecting news articles for the claims, annotators are instructed to assess the claim quality. Despite being human-generated, some claims may be ambiguous or be deemed useless if insufficient information is available online to verify or refute them.[4]

After verifying the quality of the claims, annotators search each claim on the web[5], Instagram, and Twitter to collect related news articles and posts. We call these *perspectives*. They are also instructed to provide only one unrelated perspective for each claim. Following Zarharan et al. (2019), for news article perspectives, annotators use two stance labels: the first indicates the stance of the collected news headline toward the claim (*head2claim*), and the second reflects the stance of the full article text toward the claim (*article2claim*). The stance detection task is framed as a four-way classification, requiring annotators to classify each perspective and claim as follows:[6]

- **Agree**: The perspective asserts that the claim is true without any hedging.
- **Disagree**: The perspective asserts that the claim is false without any hedging.
- **Discuss**: The perspective provides neutral information about the claim or veracity of the claim, or reports the claim without evaluating its truth.
- **Unrelated**: The claim is not addressed or reported in the perspective.

During the annotation process for *article2claim*, at least one appropriate sentence is selected as evidence from the perspective. These evidence sentences represent the minimal number of sentences necessary to justify the stance label, without needing to consider other sentences in the perspective. The collected evidence serves as a gold standard manual explanation. For samples labeled as *Unrelated*, we use the Farsi translation of "The claim is not reported in the perspective" as the explanation. Figure 2 provides examples from FARExSTANCE.

## 3.3 Quality Assurance

A group of experts or super-annotators conducted a pilot study involving six iterations of annotation and discussion. After this, the annotation team, consisting of sixteen native Farsi speakers, received training from the super-annotators. Six annotators were involved in the claim generation process, while the others focused on claim labeling.

We implemented three forms of data validation for claim labeling: inter-annotator agreement, agreement with super-annotators, and manual validation by the authors. For the news domain, two annotators labeled more than 14.5K instances, achieving a 75% agreement rate. After omitting instances without agreement, 11,648 doubly annotated instances with 100% agreement were left. A further 7,969 instances labeled by only one annotator were included, resulting in a total of 19,617 samples. We selected the instances for the validation and test sets from those with 100% annotator agreement. 10% of the social media instances were labeled by two annotators, with any conflicts excluded from the final dataset.

Annotators had agreements of 73%, 78%, and 76% with super-annotators in the news articles, tweets, and Instagram posts, respectively. Just over 1% of instances from all domains were directly checked by the authors, resulting in a 72% agreement rate between the authors and annotators.

## 3.4 Dataset Statistics

Table 5 in the Appendix illustrates the distribution of classes across training, development, and test sets. There is no overlap between the training, validation, and testing claims. FARExSTANCE includes over 26,000 instances across all domains, with 3,197 and 380 unique claims from news agencies and social media platforms, respectively. Consequently, the average Perspective Per Claim (PPC) is 6.13 for news agencies and 17.6 for social media. The minimum PPC for all domains is 1, while the maximum PPC is 122 for news agencies and 284 for social media.

## 4 Experiments

To evaluate the challenges posed by FARExSTANCE, we experiment with *XLM-RoBERTa-Large*

---

[4]116 claims were excluded from our dataset for failing to meet the required quality criteria: 36 ambiguous and 80 useless. We also revised 23 claims to correct typographical errors before using them in the dataset.

[5]We designed a search module within our web annotation tool that uses the Google search engine to search for claims and crawl relevant news articles.

[6]While Zarharan et al. (2019) defines *Agree* and *Disagree* as requiring the perspective to state the claim is true or false without any quotation, we have chosen to overlook quotations because nearly all Farsi news articles begin with a quotation, as it is a prevalent practice in Farsi news writing.

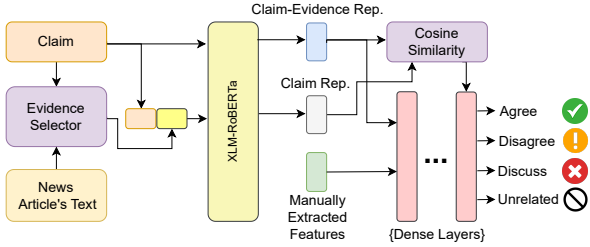

Figure 3: The baseline using *XLM-RoBERTa-Large*.

(Conneau et al., 2019) along with open- and closed-source LLMs on the news section of the dataset.

### 4.1 Baseline

Our baseline system is shown in Fig. 3. For evidence selection, we extract the $k$ sentences from the news article (or perspective) most similar to the claim (maintaining the sentence order), and use these sentences as input to *XLM-RoBERTa-Large* instead of the entire article.[7] To identify the evidence sentences, we employ the sentence transformer *MiniLM-L12*[8] along with cosine similarity. This approach addresses the sequence length limitation of *XLM-RoBERTa-Large* while also providing the two most similar sentences as explanations.

We also extract various features from the claim and news article pairs to enhance baseline performance. Since Farsi news articles typically begin with a summary and end with a conclusion relevant to the claim, we calculate the cosine similarity between the claim and the first five and last two sentences of the article as manual features. Additionally, recognizing that the stance label can depend on the presence of named entities from the claim in the article, we include the existence of up to eight named entities as a feature. For example, the stance label for **Perspective:** *Goldfish can transmit diseases to humans* against **Claim:** *Humans get skin tuberculosis from goldfish* is *Discuss*, as the specific disease (skin tuberculosis) does not appear in the perspective.[9] Lastly, inspired by prior work (Popat et al., 2019; Yang and Urbani, 2021), we calculated the cosine similarity between the claim representation and the representation of the concatenated claim and perspective. For *Agree* and *Disagree* cases, these representations should show notable (dis)similarity.

---

[7] $k$ is set to eight in our experiments.
[8] https://huggingface.co/sentence-transformers/all-MiniLM-L12-v2/discussions
[9] From *head2claim* dataset; full article omitted for brevity.

### 4.2 LLMs

We employ *Command-R-32B* (c4ai-command-r-08-2024[10]), *Llama-3.1-70B* (Dubey et al., 2024, Meta-Llama-3.1-70B), *Claude-3.5-Sonnet* (Anthropic, 2024), and *GPT-4o* (OpenAI et al., 2024) in zero-shot and few-shot scenarios, reformulating stance detection as a generation task. The models were prompted to predict the stance label and generate a corresponding explanation.

For evidence selection, we implemented the **R**etrieval **A**ugmented **G**eneration (RAG) (Lewis et al., 2020) method, which selects the two most relevant chunks of the news article to the claim using semantic chunking. The claim, a brief definition of the stance detection task, and labels, along with the selected chunks, were included in the prompt to guide the LLMs in generating both a label prediction and an explanation (see Appendix C.1 for detailed prompts). In the few-shot experiments, we also provided four instances in the prompt, with one example for each stance class from the training set, to guide the LLMs and improve their understanding of the task.

For fine-tuning, we used *Aya-23-8B* (aya-23-8B[11]), applying parameter-efficient fine-tuning methods, which aims to reduce the number of trainable parameters (Zhao et al., 2023). Specifically, we used QLoRA (Dettmers et al., 2023) and PEFT (Mangrulkar et al., 2022) for this purpose.

## 5 Results

### 5.1 Stance Classification Results

Table 1 shows the performance of different models for stance classification, evaluated using macro and weighted precision, recall, F1 and accuracy. Among open-source models in the zero-shot setting, *Command-R-32B* outperforms *Llama-3.1-70B* significantly, achieving a macro-F1 of 44.6 and an accuracy of 57.1, while *Llama-3.1-70B* struggles with a macro-F1 of 27.2 and accuracy of 26.2. The best zero-shot results come from closed-source models, with *Claude-3.5-Sonnet* and *GPT-4o* achieving macro F1 scores of 70.7 and 66.4 respectively.

Fine-tuning greatly improves performance, with *XLM-RoBERTa* leading with a macro F1 of 74.5, closely followed by *Aya-23-8B* at 72.9. In the few-shot setting, *Command-R-32B* sees significant gains, reaching a macro F1 of 56.4 and an accuracy of 68.3. The model's ability to generalize from limited examples could be due to its multilingual training on 23 languages, including Farsi. The *GPT-4o* performance in the few-shot setting is comparable with *Claude-3.5-Sonnet*, outperforming it in weighted F1 and accuracy.

---

[10] https://huggingface.co/CohereForAI/c4ai-command-r-08-2024
[11] https://huggingface.co/CohereForAI/aya-23-8B

**Human Performance on Stance Classification.**
To establish a baseline for human performance in stance classification, we randomly sampled 10% of the test set (144 out of 1,440 instances). This subset included 50 instances labeled as *Disagree*, 40 as *Agree*, 13 as *Discuss*, and 41 as *Unrelated*. To ensure unbiased evaluation, we recruited a new annotator who had no prior involvement in the original dataset collection. This annotator was provided with the guidelines and tasked with manually labeling the stance and identifying supporting evidence for each sample under time-constrained conditions. The comparison between this annotator's labels (*Human performance estimate*) and the gold-standard labels is detailed in Table 1 (last row). Human performance is on par with the best closed-source models.

## 5.2 Explanation Generation Results

**Automatic Evaluation.** To automatically evaluate the generated explanations of the models, we used ROUGE scores as well as the NLI-based reference-free coherence metrics proposed by Kotonya and Toni (2020b), as implemented by Zarharan et al. (2024). These metrics assess the logical consistency and relevance of the explanations with respect to the claims as follows:

- **Strong Global Coherence (SGC).** The explanation must fully entail the claim.
- **Weak Global Coherence (WGC).** With the exception of the instances labeled as *Disagree*, each sentence in the explanation must either entail or remain neutral toward the claim.
- **Local Coherence (LC).** No two sentences in the explanation should contradict each other.

We utilized a Farsi NLI model[12] to calculate the aforementioned coherence metrics. Table 2 shows the results on the test set. The *Similarity-based* result refers to extractive explanations comprising the two most similar sentences from the news article, identified using the *MiniLM-L12* sentence transformer. The *Human performance estimate* shows coherence metrics for the collected evidence of the new annotator (see Section 5.1). The *Human* row displays coherence metrics for our gold explanations on the test set. In terms of ROUGE-L F1 score, the *Aya-32-8B* model generates explanations that most closely align with the reference explanations. However, the few-shot approach of *GPT-4o* exhibits superior performance in SGC and WGC coherence metrics.

---

[12]https://huggingface.co/parsi-ai-nlpclass/ParsBERT-nli-FarsTail-FarSick

**Human Evaluation.** As human evaluation is essential for assessing the quality of generated text (Luo et al., 2024), the authors manually evaluated 5% of the explanations produced by each model on the test set. This subset included 21 instances labeled as *Disagree*, 22 as *Agree*, 8 as *Discuss*, and 21 as *Unrelated*, resulting in a total of 72 samples per model (864 in total). The subset shares instances with the one used in the human performance in classification (see Section 5.1), allowing for a direct comparison between the extractive explanations provided by the new annotator (referred to as *Human performance estimate*) and the gold-standard extractive explanations created by the main annotators during dataset collection. Additionally, it enables us to compare the model-generated explanations to both human-generated explanations.

The human evaluation was based on two criteria: *Suggested Class* and *Completeness*. For *Suggested Class*, we used the generated explanation (instead of the perspective in the original setting) to classify the claim into one of four labels: *Agree*, *Disagree*, *Discuss*, or *Unrelated*. If the explanation did not allow for clear classification, we assigned the label *Other*. A generated explanation is considered high quality if the annotator can accurately determine the veracity of the claim after reading it. Explanation *completeness* was assessed by selecting one of the following options: *Empty explanation*, *Includes hallucination*, *Missed details* (if the explanation failed to cover all aspects of the claim), *Incomplete generation but complete explanation* (where the model could not fit the full explanation within the maximum token limit), or *Perfect explanation*.

For the *Suggested Class* criterion, we report the macro F1 score against both the gold stance (*F1-2G*) label and the model's predicted stance label (*F1-2P*) for each instance. *F1-2G* provides insight into the quality of the generated explanation, indicating how well the explanation supports the cor-rect classification. Meanwhile, *F1-2P* reveals the consistency between the model's prediction and its explanation, highlighting whether the model's rationale aligns with its output. We develop a relaxed scoring metric called the **O**verall **E**xplanation **S**core (*OES*) to evaluate the best model for explanation generation. This score is derived from averaging four key metrics: *F1-2G*, *F1-2P*, the **P**ercentage of **P**erfect **E**xplanations (*PEP*), and the percentage of **I**ncomplete **G**eneration but **C**omplete **E**xplanation (*IGCE*). Since we considered PEP and IGCE as collectively indicative of a sufficiently good explanation, we divided their combined sum by three to obtain the final score. Table 3 shows the results.[13]

| Setting | Method Model | Macro | | | Weighted | | | Acc |
|---|---|---|---|---|---|---|---|---|
| | | P | R | F1 | P | R | F1 | |
| Zero | Claude-3.5-Sonnet | 71.8 | 70.5 | **70.7** | 77.7 | 75.3 | 76.0 | 77.8 |
| | GPT-4o | 68.3 | 67.1 | 66.4 | 74.9 | 76.7 | 74.8 | 76.7 |
| | Command-R-32B | 43.9 | 47.8 | **44.6** | 51.8 | 57.1 | 52.9 | 57.1 |
| | Llama-3.1-70B | 33.9 | 28.1 | 27.2 | 37.9 | 26.2 | 28.7 | 26.2 |
| Few | Claude-3.5-Sonnet | 72.8 | 72.0 | **71.5** | 79.6 | 76.8 | 77.4 | 76.8 |
| | GPT-4o | 73.4 | 70.9 | 70.8 | 77.9 | 79.8 | 78.1 | **79.8** |
| | Command-R-32B | 58.8 | 58.7 | **56.4** | 66.1 | 68.3 | 65.2 | 68.3 |
| | Llama-3.1-70B | 32.9 | 27.3 | 26.7 | 36.9 | 25.7 | 28.3 | 25.7 |
| PEFT | Aya-23-8B | 73.1 | 72.8 | 72.9 | 78.2 | 78.5 | **78.2** | 78.3 |
| RoBERTa | XLM-RoBERTa | 75.2 | 74.2 | **74.5** | 78.1 | 78.7 | **78.2** | 78.7 |
| *Majority baseline* | | 7.7 | 25.0 | 11.7 | 9.4 | 30.6 | 14.4 | 30.6 |
| *Human performance estimate* | | 68.3 | 68.2 | 68.0 | 79.9 | 79.2 | 79.2 | 79.2 |

Table 1: Performance of the stance classification models on the test set, reported using macro and weighted scores. Acc, P, and R represent Accuracy, Precision, and Recall, respectively.

| | Model | RL | SGC | WGC | LC |
|---|---|---|---|---|---|
| Zero | Claude-3.5-Sonnet | 7.2 | 1.2 | 71.5 | 60.1 |
| | GPT-4o | 6.0 | 31.4 | 75.3 | 80.3 |
| | Command-R-32B | 4.4 | 10.1 | 80.7 | **92.4** |
| | Llama-3.1-70B | 5.8 | 9.5 | 70.6 | 64.4 |
| Few | Claude-3.5-Sonnet | 11.2 | 1.2 | 63.6 | 28.2 |
| | GPT-4o | 8.8 | 21.5 | **82.6** | 87.8 |
| | Command-R-32B | 13.8 | 2.1 | 80.1 | 74.0 |
| | Llama-3.1-70B | 5.4 | 9.7 | 69.1 | 64.4 |
| Aya-23-8B (PEFT) | | **17.5** | 5.2 | 79.8 | 80.5 |
| Similarity-based | | 9.6 | 1.9 | 80.8 | 85.1 |
| *Human PE.* | | 28.1 | 3.5 | 79.9 | 84.0 |
| *Human* | | - | 8.5 | 77.8 | 74.3 |

Table 2: ROUGE-L F1 scores, and NLI metrics on the test set for explanation generation. **Human PE** stands for *Human performance estimate*.

| | Model | F1s | PEP | IGCE | OES |
|---|---|---|---|---|---|
| Zero | Claude-3.5-Sonnet | 77.8 / 85.2 | 11.1 | **77.8** | 84.0 |
| | GPT-4o | 73.7 / 85.8 | **93.1** | 0.0 | **84.2** |
| | Command-R-32B | 50.5 / 47.1 | 47.2 | 0.0 | 48.3 |
| | Llama-3.1-70B | 26.4 / 36.1 | 31.9 | 13.9 | 36.1 |
| Few | Claude-3.5-Sonnet | 82.9 / 87.4 | 52.8 | 40.3 | **87.8** |
| | GPT-4o | 75.4 / 75.3 | 86.1 | 1.4 | 79.4 |
| | Command-R-32B | 63.8 / 61.4 | 48.6 | 33.3 | 69.0 |
| | Llama-3.1-70B | 35.2 / 44.9 | 19.4 | 12.5 | 37.3 |
| Aya-23-8B (PEFT) | | 74.2 / 61.5 | 66.7 | 12.5 | **71.6** |
| *Similarity-based* | | 62.2 / 53.0 | 30.6 | 2.8 | 49.5 |
| *Human PE.* | | 71.9 / 66.0 | 69.4 | 0.0 | 69.1 |
| *Human* | | **87.6 / 87.6** | 90.3 | 0.0 | **88.5** |

Table 3: Human evaluation results: The **F1s** denotes *F1-2G* and *F1-2P* respectively. **Human PE** stands for *Human performance estimate*

The results highlight that the gold-standard human explanations outperform all LLMs with the highest *F1-2P* score of 87.6 and an *OES* of 88.5. The few-shot *Claude-3.5-Sonnet*, despite its limitations in handling the maximum number of new tokens, achieves the highest performance among all LLMs, with an *OES* of 87.8, followed by the zero-shot *GPT-4o*. A comparison between zero-shot and few-shot settings reveals that all models, except *GPT-4o*, benefit significantly from the examples provided in the prompt. *GPT-4o* excels in the *PEP* with 93.1 and its *IGCE* score is 0. Models like *Aya-23-8B* and few-shot *Command-R-32B* approaches show moderate results, with an *OES* of 71.6 and 69.0 respectively. The *Human PE* is relatively low, primarily due to the new annotator missing important details in 30.6% of instances when selecting relevant sentences as explanations (see Table 7). This oversight significantly impacted performance, particularly for *Discuss* instances. See Fig. 8 in Appendix E for a comparison of the generated explanations from all models for one example.

---
[13]See Table 6 in the Appendix for full results.

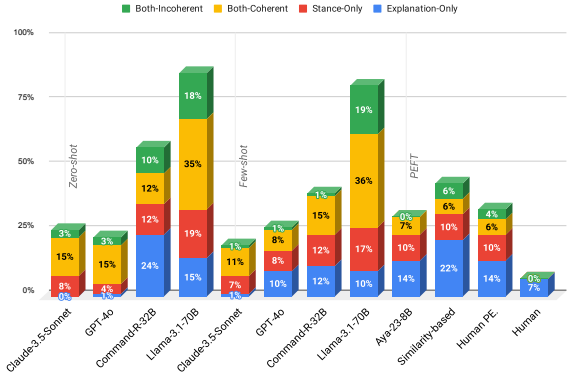

Figure 4: The percentage distribution of error types across various models and settings.

## 6 Analysis

### 6.1 Exploring Dataset Bias

To explore potential biases in our dataset, we conduct experiments using two distinct models. In the first model, we use only the claim as input to detect biases related specifically to the claims. In the second, we provide only the perspective as input to uncover biases within the news article text. For both models, we employed *XLM-RoBERTa-Large* and trained it separately using only the claims for one model and only the contexts (news article content) for the other. After training, the models simply default to predicting the majority class (*Agree* class), leading to performance equivalent to that of a majority-class baseline (see Table 1) and demonstrating a failure to learn meaningful patterns.

### 6.2 Error Analysis

**Common Stance Prediction Errors.** We identified all instances in the test set that were misclassified by all models. The results show 28 instances with gold labels as *Discuss*, 12 as *Disagree*, 8 as *Unrelated*, and 5 as *Agree*. This indicates that instances with the *Discuss* label pose the greatest challenge, while those labeled as *Agree* are the easiest to classify correctly. In the randomly sampled 10% of the test set which was used to compare *Human performance estimate* (see Section 5.1), the new annotator also struggled most with classifying 11 *Discuss* instances, followed by 7 *Disagree* and *Unrelated* instances, and only 3 misclassifications of *Agree* instances. These findings align with the confusion matrices of the models, which reveal that most struggled to accurately classify the *Discuss* category. As shown in Table 6 in the Appendix, our human evaluation confirms that all models, with the exception of *Similarity-based*, face similar difficulties in classifying the *Discuss* category.

**Analysis of Human Evaluation Results.** Focusing on the instances used in our human evaluation (72 per model, see Section 5.2), we classified errors into four categories: *Explanation-Only*, *Stance-Only*, *Both-Coherent*, and *Both-Incoherent*. An *Explanation-Only* error occurs when the model predicts the correct stance but the explanation does not match. A *Stance-Only* error occurs when the model predicts an incorrect stance but its explanation matches the gold stance. The *Both* error type refers to cases where both the generated explanation and the predicted stance do not match the gold stance. In a *Both-Coherent* error, the model generates an explanation consistent with its prediction, i.e. the *Suggested Class* matches the predicted class. In contrast, a *Both-Incoherent* error occurs when the *Suggested Class* differs from both the predicted class and the gold class.

Fig. 4 illustrates the percentage distribution of these error types across different models and settings. The overall failure rate for all models except *GPT-4o* decreased in the few-shot setting compared to the zero-shot setting. Since annotators were tasked with selecting specific sentences as evidence (extractive explanations), the last bar in the chart (Human) indicates that, for a small percentage of examples, not all relevant sentences were selected. This omission led to *Explanation-Only* errors. This failure type is illustrated in the first and second examples in Fig. 5 and typically happens when the model fails to include certain details of the claim (highlighted in pink) in its generated explanation. A *Stance-Only* error typically arises when the model fails to adequately understand the claim and its context, leading to an incorrect prediction accompanied by a context summary as the explanation (see the third example in Fig. 5). *Both*

| Failed Examples |
|---|
| **Example 1: Aya-23-8B (PEFT) - Explanation-Only** |
| ادعا: جو بایدن، رئیس جمهوری ایالات متحده، به ارزش‌های جهانی از جمله برابری ادای احترام کرد<br>**Claim:** Joe Biden, the President of the United States, paid tribute to universal values, including equality.<br>توضیح تولید شده: با آغاز ماه محرم جو بایدن رئیس جمهوری آمریکا پیامی به این مناسبت در صفحه توئیتری خود منتشر کرد . بایدن به مناسبت ماه محرم پیام داد ۱۸ مرداد ۱۴۰۰ ، ۲۳ : ۴۰ : ۱۲۴۰۰۰۵۱۸۰۸۹ نظر , ۰ در صف انتشار ۰ ۵ تکراری یا غیرقابل انتشار با آغاز ماه محرم جو بایدن رئیس جمهوری آمریکا پیامی به این مناسبت در صفحه توئیتری خود منتشر کرد . بایدن نوشت : من و جیل ( همسرم ) برای همه کسانی که سال نو اسلامی ( اول محرم الحرام ) و یاد فداک...<br>**Generated Explanation (Incomplete Generation):** With the beginning of the month of Muharram, Joe Biden, the President of the United States, published a message on this occasion on his Twitter page. Biden wrote: Jill and I [my wife] send our wishes to all those observing the Islamic New Year (the first of Muharram) and honoring the memory of sac...<br>**Suggested Class:** Unrelated  **Predicted Class:** Agree  **Gold Class:** Agree |
| **Example 2: GPT-4o (Zero) - Explanation-Only** |
| ادعا: فرمانده مرزبانی نیروی انتظامی اظهارکرد: با اقدام موفق سپاه پاسداران ۲ تن از مرزبانان گروگان گرفته شده توسط گروهک ضد انقلاب جیش الظلم، به وطن بازگشتند.<br>**Claim:** The commander of the Border Guard of the Law Enforcement Force stated: With the successful operation of the IRGC, two border guards who were taken hostage by the counter-revolutionary group Jaish al-Zolm have returned to the homeland.<br>توضیح تولید شده: این متن به وضوح تأیید می‌کند که دو مرزبان ربوده شده توسط گروهک تروریستی جیش الظلم با اقدام موفق نیروی زمینی سپاه پاسداران به وطن بازگشتند.<br>**Generated Explanation:** This text clearly confirms that the two border guards abducted by the terrorist group Jaish al-Zolm have returned to the homeland through the successful operation of the IRGC Ground Forces.<br>**Suggested Class:** Discuss  **Predicted Class:** Agree  **Gold Class:** Agree |
| **Example 3: Llama-3.1-70B (Zero) - Stance-Only** |
| ادعا: یک هیات از دانشمندان سازمان ملل طی یک هشدار در گزارشی، نزدیک شدن گرمایش زمین به خط قرمز را اعلام کردند.<br>**Claim:** A panel of United Nations scientists, in a warning issued in a report, announced that global warming is approaching a critical red line.<br>توضیح تولید شده: این متن به صورت خبری در مورد هشدار دانشمندان سازمان ملل در مورد گرمایش زمین و نزدیک شدن آن به خط قرمز گزارش شده است. در متن آمده است که تغییرات آب و هوا شدید و سریع شده است، اما در متن اشاره ای به این که این تغییرات ناشی از گرمایش زمین است، نشده است. در متن آمده است که دی اکسید کربن در جو زمین به بالاترین حد در دو میلیون سال اخیر رسیده است، اما در متن اشاره ای به این که این افزایش ناشی از گرمایش زمین است، نشده است. در متن آمده است که گازهای گلخانه ای ناشی از سوخت های فسیلی عامل اصلی این تغییرات...<br>**Generated Explanation (Incomplete Generation):** This text is presented as a news report about UN scientists warning of global warming and its approach to the red line. It mentions that climate changes have become severe and rapid, but the text does not indicate that these changes are due to global warming. It states that carbon dioxide in Earth's atmosphere has reached its highest level in two million years, but the text does not mention that this increase is caused by global warming. It notes that greenhouse gases from fossil fuels are the main drivers of these changes...<br>**Suggested Class:** Agree  **Predicted Class:** Discuss  **Gold Class:** Agree |
| **Example 4: Claude-3.5-Sonnet (Zero) - Both-Coherent** |
| ادعا: بنابر اعلام باشگاه فوتبال بارسلونا، لیونل مسی به دلیل موانع ساختاری از باشگاه بارسلونا جدا می شود.<br>**Claim:** According to the announcement by FC Barcelona, Lionel Messi will leave the club due to structural obstacles.<br>توضیح تولید شده به فارسی: توضیح به فارسی: بر اساس متن ارائه شده، این ادعا صحیح است. در متن آمده است که "چند شب پیش بود که باشگاه بارسلونا در خبری غیر قابل باور برای هوادارانش اعلام کرد که کار لیونل مسی در بلوگرانا به پایان رسیده و او فصل آینده با پیراهن تیم دیگری به مید...<br>**Generated Explanation (Incomplete Generation):** Explanation in Farsi: Based on the provided text, this claim is correct. The text states that "A few nights ago, Barcelona Football Club, in an unbelievable announcement for its fans, declared that Lionel Messi's time at Blaugrana had come to an end and that next season he would appear in the jersey of another team to ...<br>**Suggested Class:** Agree  **Predicted Class:** Agree  **Gold Class:** Discuss |
| **Example 5: Command-R-32B (Zero) - Both-Incoherent** |
| ادعا: بانک جهانی: پس از تسلط طالبان بر افغانستان کمک های مالی به این کشور متوقف شده است.<br>**Claim:** The World Bank: Financial aid to Afghanistan has been halted following the Taliban's takeover of the country.<br>توضیح تولید شده: بانک جهانی در بیانیه ای از توقف اختصاص کمک های مالی به افغانستان خبر داده است. سخنگوی بانک جهانی اعلام کرده که این نهاد مستقر در واشنگتن با دقت اوضاع افغانستان را زیر نظر دارد. وی ادامه داد: ما به شدت نگران اوضاع در افغانستان و اثرات آن بر چشم اندازهای توسعه و پیشرفت این کشور علی الخصوص برای زنان هستیم. ادعا می شود که بانک جهانی به رایزنی با جامعه بین الملل و شرکای توسعه (در افغانستان) در مورد وضعیت ادامه خواهد داد و در حال بررسی راه هایی برای حفظ مشارکت جهت حفظ دستاوردهای توسعه ای که با زحمت به دست آمده است. بانک جهانی...<br>**Generated Explanation (Incomplete Generation):** The World Bank, in a statement, announced the suspension of financial aid to Afghanistan. A spokesperson for the World Bank stated that the Washington-based organization is closely monitoring the situation in Afghanistan. He continued: "We are deeply concerned about the situation in Afghanistan and its impact on the country's development prospects, particularly for women." It is claimed that the World Bank will continue consulting with the international community and development partners (in Afghanistan) regarding the situation and is exploring ways to maintain engagement to preserve the hard-earned development gains. The World Bank...<br>**Suggested Class:** Discuss  **Predicted Class:** Unrelated  **Gold Class:** Agree |

Figure 5: Examples of failure cases by category. Missed details are highlighted in pink, while incomplete generated explanations are marked in light red for clarity.

errors often occur when the model overlooks critical details of the claim during stance label classification and explanation generation. The last two examples in Fig. 5 illustrate *Both-Coherent* and *Both-Incoherent* errors, respectively, with the missed details highlighted in pink.

Manual inspection of the generated explanations reveals that all models, particularly open-source LLMs, struggle with capturing fine details. This ranges from 4.2% of instances for *Claude-3.5-Sonnet* in a few-shot setting, up to 50.0% for *Command-R-32B* in a zero-shot setting (see Table 6 for more details). There is significant room for improvement in delivering more fine-grained and accurate explanations for Farsi stance detection.

## 7 Conclusion

We introduced a new dataset for Farsi explainable stance detection, FAREXSTANCE, and conducted extensive experiments to establish baseline performance using a variety of multilingual open-source small and large language models, retrieval-augmented generation, and parameter-efficient fine-tuning approaches. We also provided insights into the strengths and limitations of these approaches using both automatic and human evaluations. Our dataset, manually curated with labeled instances and supporting evidence, has been made publicly available to foster further research in this area, e.g. experiments with the social media perspectives.

## 8 Limitations

We acknowledge the following limitations in our study:

1. Due to computational constraints, fine-tuning was only feasible for the *Aya-23-8B* model. We were unable to fine-tune *Command-R-32B* and *Llama-3.1-70B* due to their larger size and our hardware limitations.
2. Among closed-source Large Language Models (LLMs), we limited our exploration to *Claude-3.5-Sonnet* and *GPT-4o*. Budget constraints prevented us from evaluating additional closed-source LLMs, which typically incur significant usage costs.
3. Our experiments were conducted exclusively on the Farsi language using our proposed dataset. The generalizability of our findings to other languages or datasets remains to be investigated.
4. The social media domain and the implementation of *head2claim* models, which could be particularly valuable for stance detection in social media perspectives, were not explored in our experiments and are left for future work.

## Acknowledgements

The authors sincerely thank Dadmatech for their invaluable support and collaboration, which made the collection of the dataset for this research possible. Their contribution has been instrumental in enabling the progress and completion of this work. The authors would like to thank Babak Behkam Kia and the annotation team for their significant contribution. The experiments of this research are also supported by Science Foundation Ireland (SFI) through the SFI Frontiers for the Future programme (19/FFP/6942) and the ADAPT Centre for Digital Content Technology, which is funded under the SFI Research Centres Programme (Grant 13/RC/2106) and is co-funded under the European Regional Development. We thank the reviewers for their insightful and helpful comments.

## Ethical Considerations

This research was conducted in adherence to established ethical guidelines and platform policies, with an emphasis on user privacy and data protection. We implemented data anonymization and minimization techniques to safeguard user information, ensuring that only essential data was collected and securely stored. All data was used exclusively for academic purposes and collected in strict compliance with the non-commercial use policies of Twitter (now X) and Instagram.

Data access was facilitated through official APIs, and to protect user content, only post identifiers (post IDs) were included in the released dataset, rather than full content. This approach minimizes the risk of exposing sensitive user information while maintaining the utility of the dataset for research purposes.

## A Dataset Comparison

Table 4 provides a comprehensive comparison of our dataset with existing datasets in the domains of stance detection, fact-checking, and fake news detection.

## B Dataset Collection Details

### B.1 Claim Mutation

During the claim mutation phase, by following Thorne et al. (2018) and Zarharan et al. (2021) we define six types of mutations: paraphrasing, negation, substitution of an entity with a similar or dissimilar one, and altering the claim to make it either more general or more specific. The latter category is further divided into two subtypes: (1) adding one or more words or phrases to the original claim based on the annotator's own knowledge, and (2) incorporating information from the summary. Annotators are tasked with generating one to three mutated claims for each mutation type based on the following guidelines:

1. **Paraphrasing**: Each generated claim is rephrased in a different way while retaining the same meaning and concept. Try to replace words with their synonyms as much as possible, and you can also rearrange the position of words in the sentence. **Notes:** The mutated claim must be entailed by the original claim, and the original claim must also be entailed by the mutated claim.

2. **Negation**: Modify the claim in a way that contradicts it and effectively opposes the meaning of the original claim. **Notes:** Avoid negating the claim merely by adding words like "not" at the beginning of verbs. Instead, strive to achieve this by using antonyms and altering the concept of the sentence.

3. **Substitution with a similar entity**: Replace one of the entities (elements) of the claim with a word from its corresponding category. **Notes:** The replacement word should not convey exactly the same meaning, as this type of mutation pertains to paraphrasing. The modified claim should not allow us to infer the original claim. Furthermore, the mutated claim should not contradict the original claim, as this would fall under negation.

4. **Substitution with a dissimilar entity**: This type of mutation is similar to the previous one, but the selected entity is replaced with something entirely different and unrelated. **Notes:** The modified claim should not allow us to infer the original claim. Additionally, the generated claim should not contradict the original claim.

5. **More specific claim:** The claim should be expressed in a more detailed manner by including additional information. This should be done twice: once by incorporating information from the news summary and once by relying on personal knowledge. **Notes:** The mutated claim must entail the original claim but should not convey exactly the same meaning, as this would fall under paraphrasing.

6. **More general claim**: In contrast to the previous rule, this guideline requires the claim to be expressed in a more general manner, containing less specific information. **Notes:** The original claim should entail the mutated claim, but the mutated claim must not convey exactly the same meaning as the original, as this would fall under paraphrasing. Additionally, generalization should not be achieved merely by omitting words.

| Dataset | Type | Language | Domain | Labels | Size | Explanation | Explanation Type |
|---|---|---|---|---|---|---|---|
| Fake News Challenge (FNC-1) | Target-specific | English | News articles | Agree, Disagree, Discuss, Unrelated | 49.9k | No | - |
| COVID-CQ (Mutlu et al., 2020) | Target-specific | English | Tweet | Favor, Against, Neutral | 14.3k | No | - |
| P-Stance (Li et al., 2021) | Target-specific | English | Tweet | Favor, Against, Neutral | 21.5k | No | - |
| SemEval-2016 (Mohammad et al., 2016) | Multi-target | English | Tweet | Favor, Against, None | 4.8k | No | - |
| Stance-hof (Grimminger and Klinger, 2021) | Multi-target | English | Tweet | Favor, Against, Neither, Mixed, Neutral | 3k | No | - |
| COVMis-Stance (Hou et al., 2022) | Multi-target | English | Tweet | Favor, Against, Neither | 2.6k | No | - |
| VaccineLies (Weinzierl and Harabagiu, 2022) | Multi-target | English | Tweet | Accept, Reject, No Stance | 14.6k | No | - |
| WT-WT (Conforti et al., 2020) | Multi-target | English | Tweet | Support, Refute, Comment, Unrelated | 51.2k | No | - |
| COVID-19-Stance (Glandt et al., 2021) | Multi-target | English | Tweet | In-favor, Against, Neither | 6.1k | No | - |
| VAST (Allaway and McKeown, 2020) | Multi-target | English | ARC Corpus | Pro, Con, Neutral | 23.5k | No | - |
| ToxiChat (Baheti et al., 2021) | Claim-based | English | Reddit conversations | Agree, Disagree, Neutral | 2k | No | - |
| Monant Medical Misinformation (Srba et al., 2022) | Claim-based | English | News Article | supporting, contradicting, neutral, Can't tell | 0.3k | No | - |
| FEVER (Thorne et al., 2018) | Claim-based | English | Wikipedia articles | Support, Refute, NotEnoughInfo | 185.4k | No | - |
| HOVER (Jiang et al., 2020) | Claim-based | English | Wikipedia articles | Support, Refute, NotEnoughInfo | 26.1k | No | - |
| e-FEVER (Stammbach and Ash, 2020) | Claim-based | English | Wikipedia articles | Support, Refute, NotEnoughInfo | - | Yes | Machine Generated |
| EX-FEVER (Ma et al., 2024) | Claim-based | English | Wikipedia articles | Support, Refute, NotEnoughInfo | 60k | Yes | Human Generated |
| LIAR (Wang, 2017) | Claim-based | English | short statements from POLITIFACT.COM's API | Pants-Fire, False, Barely-True, Half-True, Mostly-True, True | 12.8k | No | - |
| LIAR-PLUS (Alhindi et al., 2018) | Claim-based | English | short statements from POLITIFACT.COM's API | Pants-Fire, False, Barely-True, Half-True, Mostly-True, True | 12.8k | Yes | Human Generated |
| PUBHEALTH (Kotonya and Toni, 2020a) | Claim-based | English | fact checking websites and news/news review websites | True, False, Unproven, Mixture | 11.8k | Yes | Human Generated |
| CIC-ES (Zotova et al., 2020) | Target-specific | Spanish | Tweet | Favor, Against, Neutral | 10k | No | - |
| CIC-CA (Zotova et al., 2020) | Target-specific | Catalan | Tweet | Favor, Against, Neutral | 10k | No | - |
| Twitter Stance Election 2020 (Kawintiranon and Singh, 2021) | Target-specific | English | Tweet | Against, Favor, None | 2.5k | No | - |
| Conversational Stance Detection (Li et al., 2022) | Target-specific | Cantonese | posts/comments in conversation threads in social media | Favor, Against, Neither | 5.8k | No | - |
| ExaASC (Jaziriyan et al., 2021) | Multi-target | Arabic | Tweet | Favor, Against, None | 9.5k | No | - |
| x-stance (Vamvas and Sennrich, 2020) | Multi-target | German, French, Italian, English | comments in Smartvote website | Favor, Against | 67k | No | - |
| RuStance (Lozhnikov et al., 2018) | Claim-based | Russian | Tweet and news | Support, Deny, Query, Comment | 0.9k | No | - |
| French Tweet Corpus (Evrard et al., 2020) | Claim-based | French | Tweet | Support, Deny, Query, Comment, Ignore | 5.8k | No | - |
| Persian Stance Classification (Zarharan et al., 2019) | Claim-based | Farsi | News Article | Agree, Disagree, Discuss, Unrelated | 2.1k | No | - |
| ParsFEVER (Zarharan et al., 2021) | Claim-based | Farsi | Wikipedia pages | Support, Refute, NotEnoughInfo | 23k | No | - |
| Persian Tweets Stance Detection (Bokaei et al., 2022) | Claim-based | Farsi | Tweet | In Favor of, Against, Neutral | 3.8k | No | - |
| **FarExStance (Our dataset)** | **Claim-based** | **Farsi** | **Tweet, Instagram post, News Headline and Article** | **Agree, Disagree, Discuss, Unrelated** | **26.3k** | **Yes** | **Human Generated** |

Table 4: Comparison of stance detection datasets. **Dataset** specifies the name of the dataset along with a citation. **Type** indicates the classification approach used, such as Claim-based, Target-specific, or Multi-target. **Language** specifies the language of the dataset. **Domain** specifies the source of the data, for instance, Tweets or News articles. **Labels** lists the stance categories employed, e.g., Favor, Against, Neutral. **Size** quantifies the total number of instances in the dataset. **Explanation** suggests whether the dataset includes justifications for the stance labels. **Explanation Type** classifies the nature of the explanations, such as Human Generated or Machine Generated.

Some other general notes for the mutation phase are as follows:

- All generated claims, across different mutation types, should be centered around the title and the provided summary of the news.
- In the Similar and Dissimilar sections, when selecting an entity, it is permissible to modify only one entity at a time.
- In the Similar and Dissimilar sections, adjec-

tives and adverbs should not be treated as entities.

- In the Similar section, when selecting occupations, organizations, or similar entities, they must belong to a closely related set.

- Regarding positions and titles as entities: If the modified claim contradicts the original claim (or vice versa), it is incorrect to select a position as the entity.

Dataset statistics are shown in Table 5. Screenshots of our annotation tool are shown in Fig. 6.

## C Experimental Details

### C.1 Prompts

To optimize model performance, we conducted a systematic prompt engineering process. We tested a diverse set of prompts for each LLM on a subset of the development set. The efficacy of each prompt was evaluated through a manual assessment. Based on these evaluations, we identified the most effective prompt for each model. The final prompts used in our experiments are as follows:

**Claude-3.5-Sonnet & GPT-4o:**

```
System Message: You are a helpful
assistant that predicts the stance of a
context against a claim and explains the
reason for your prediction by considering
the context.  Instructions: {Few-shot
Samples} Categorize the stance of the
following context against the claim as:
- Agree: Context unequivocally supports
the claim's truth.
- Disagree: Context unequivocally refutes
the claim's truth.
-  Discuss:   Context  offers  neutral
information or reports the claim without
evaluating its veracity, or context missed
some details in the claim.
- Unrelated: Claim is not addressed in the
context.

Output format should be:
[Category]
[Explanation in Farsi]
Provide  only  the  category  and  Farsi
explanation based only on the context.
Think step-by-step before you write the
response.

User Message: Context:{Context}
Claim:{Claim}
```

**Command-R-32B:**

```
###Instruction:
Use the Task below and the Input given
to write the Response, which is a stance
label prediction that can solve the Task.

###Task:
Categorize the stance of the following
context against the claim as:
- Agree:  Context unequivocally supports
the claim's truth.
- Disagree: Context unequivocally refutes
the claim's truth.
-  Discuss:   Context  offers  neutral
information or reports the claim without
evaluating its veracity, or context missed
some details in the claim.
- Unrelated: Claim is not addressed in the
context.

Output format should be:
[Category]
[Explanation in Farsi]
Provide  only  the  category  and  Farsi
explanation based only on the context. No
additional text!!!!.

###Input:
Context: {context}
Claim: {claim}

{discriminant} Response:
```

**Llama-3.1-70B:**

```
###Instruction:
Use the Task below and the Input given
to write the Response, which is a stance
label prediction that can solve the Task.

###Task:
Categorize the stance of the following
context against the claim as:
- Agree:  Context unequivocally supports
the claim's truth.
- Disagree: Context unequivocally refutes
the claim's truth.
-  Discuss:   Context  offers  neutral
information or reports the claim without
evaluating its veracity, or context missed
some details in the claim.
- Unrelated: Claim is not addressed in the
context.

Output format should be:
[Category]
[Explanation in Farsi]

Provide  only  the  category  and  Farsi
explanation based only on the context. No
additional text!!!!.  Think step-by-step
before you write the response.

###Input:
Context: {context}
Claim: {claim}

{discriminant} Response:
```

| Domain | Data Split | Disagree | Agree | Discuss | Unrelated | Total |
|---|---|---|---|---|---|---|
| News Agencies | Train | 4,886 | 5,376 | 2,147 | 4,504 | 16,913 |
|  | Val | 327 | 393 | 161 | 383 | 1,264 |
|  | Test | 430 | 441 | 166 | 403 | 1,440 |
| Social Media | Train | 892 | 1,208 | 1,037 | 1,962 | 5,099 |
|  | Val | 29 | 182 | 134 | 243 | 588 |
|  | Test | 369 | 214 | 129 | 291 | 1,003 |

Table 5: The distribution of classes across training, validation, and test sets for all domains.

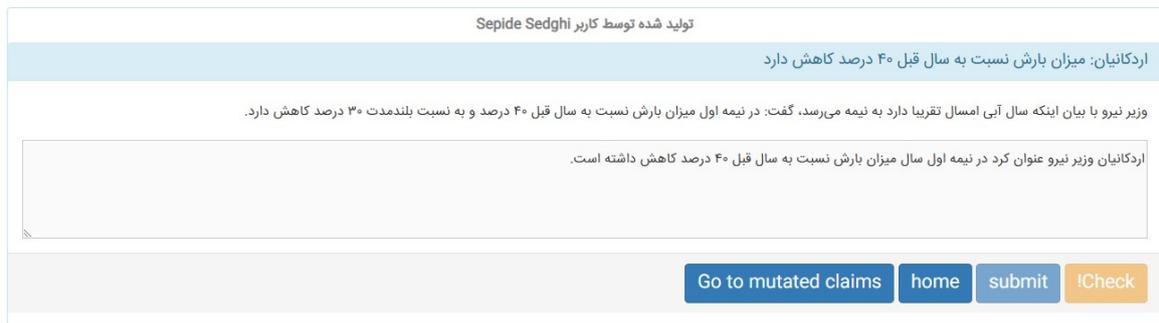

(a) Claim generation phase.

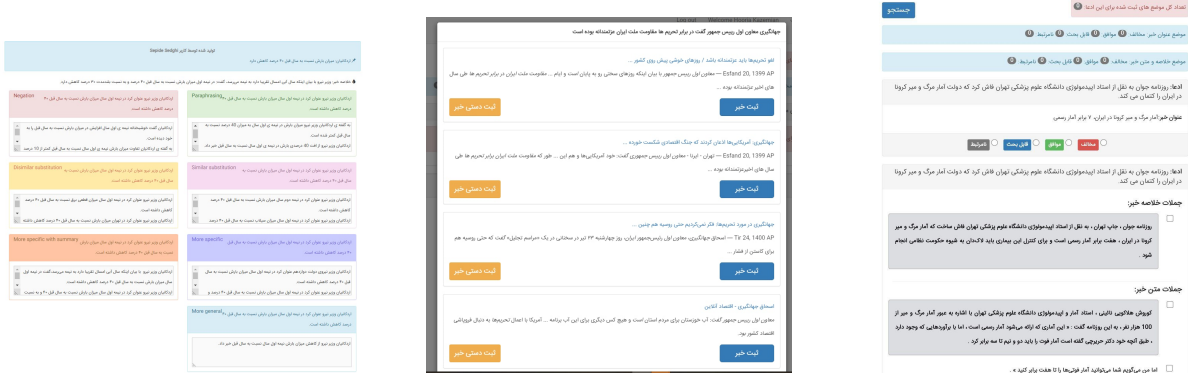

(b) Claim mutation phase.     (c) Collecting news articles.     (d) Label generation phase.

Figure 6: The screenshots of our annotation tool for collecting the FarExStance dataset.

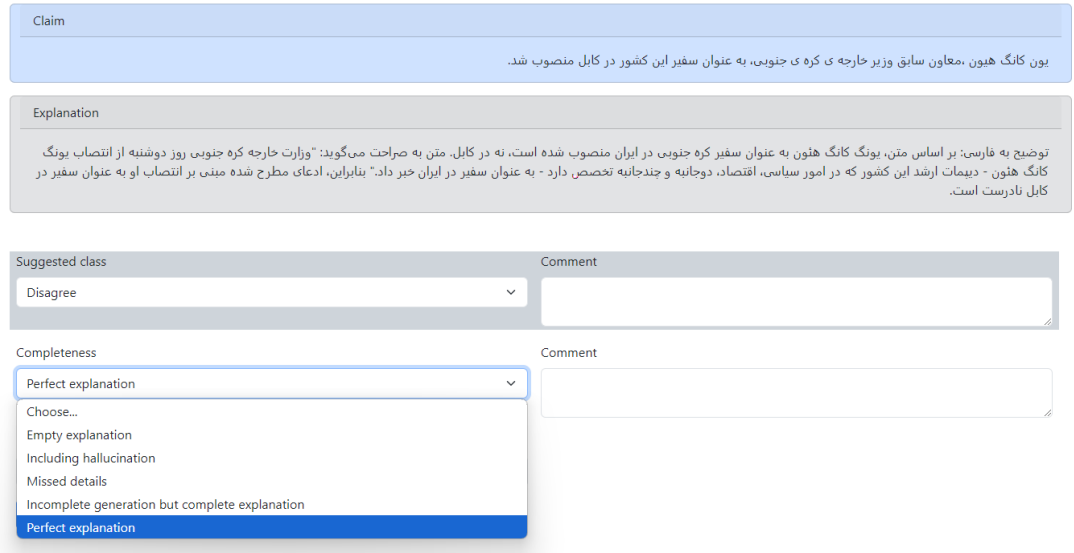

Figure 7: A screenshot of our annotation tool for human evaluation.

## C.2 Experiment Setting Details

We conducted zero-shot and few-shot experiments using default hyperparameters for all selected LLMs. Due to computational constraints, we quantized *Command-R-32B* and *Llama-3.1-70B* to 4-bit precision for our in-context learning experiments. Across all LLMs, we limited the maximum generation length to 150 tokens. We used also *MiniLM-L12*[14] as the sentence transformer in RAG using semantic chunking

For *Aya-23-8B*, we implemented parameter-efficient fine-tuning using 8-bit quantization. The fine-tuning process utilized the AdamW optimizer (specifically, paged_adamw_32bit) with a learning rate of 2e-4. We explored various hyperparameter configurations, selecting the optimal values based on validation set performance. In our QLoRA settings, we empirically determined the best values for *r* and *alpha* to be 16, and set *lora_dropout* to 0.5. Following QLoRA's default configuration, we set *bias* to 'none' and *task_type* to 'CAUSAL_LM'.

## D Detailed Results

Results of our human evaluation of the generated explanations are shown in Tables 6 and 7. Figure 7 displays the screenshot of the annotation tool used for our human evaluation.

## E Analysis

Fig. 8 compares the generated explanations from all models for a specific example. In this instance, few-shot *Claude-3.5-Sonnet* and *GPT-4o* produced the best explanations, as their outputs were complete and aligned with their predicted stance. In contrast, few-shot *Llama-3.1-70B* and *Command-R-32B* provided incomplete generations but complete explanations, though their predicted stance classes were inaccurate. Notably, when examining the *Human PE*, it is evident that the annotator did not select all relevant sentences as part of the extractive explanation, even though the predicted class matched the gold class.

---

[14]https://huggingface.co/sentence-transformers/all-MiniLM-L12-v2/discussions

| | Model | Class | F1-2G | F1-2P | PEP | IGCE | OES | CEP |
|---|---|---|---|---|---|---|---|---|
| **Zero-shot** | Claude-3.5-Sonnet | All | 77.8 | 85.2 | 11.1 | 77.8 | 84.0 | 9.7 / 1.4 / 0.0 |
| | | Agree | 95.2 | 100.0 | 13.6 | 81.8 | 96.9 | 4.5 / 0.0 / 0.0 |
| | | Disagree | 100.0 | 97.6 | 23.8 | 71.4 | 97.6 | 4.8 / 0.0 / 0.0 |
| | | Discuss | 66.7 | 66.7 | 0.0 | 50.0 | 61.1 | 50.0 / 0.0 / 0.0 |
| | | Unrelated | 80.0 | 88.0 | 0.0 | 90.5 | 86.2 | 4.8 / 4.8 / 0.0 |
| | GPT-4o | All | 73.7 | 85.8 | 93.1 | 0.0 | **84.2** | 6.9 / 0.0 / 0.0 |
| | | Agree | 90.0 | 97.3 | 95.5 | 0.0 | 94.3 | 4.5 / 0.0 / 0.0 |
| | | Disagree | 100.0 | 100.0 | 95.2 | 0.0 | 98.4 | 4.8 / 0.0 / 0.0 |
| | | Discuss | 54.5 | 80.0 | 87.5 | 0.0 | 74.0 | 12.5 / 0.0 / 0.0 |
| | | Unrelated | 86.5 | 93.3 | 90.5 | 0.0 | 90.1 | 9.5 / 0.0 / 0.0 |
| | Command-R-32B | All | 50.5 | 47.1 | 47.2 | 0.0 | 48.3 | 50.0 / 2.8 / 0.0 |
| | | Agree | 62.5 | 62.1 | 45.5 | 0.0 | 56.7 | 54.5 / 0.0 / 0.0 |
| | | Disagree | 64.5 | 66.7 | 42.9 | 0.0 | 58.0 | 52.4 / 4.8 / 0.0 |
| | | Discuss | 40.0 | 0.0 | 12.5 | 0.0 | 17.5 | 87.5 / 0.0 / 0.0 |
| | | Unrelated | 89.5 | 83.9 | 66.7 | 0.0 | 80.0 | 28.6 / 4.8 / 0.0 |
| | Llama-3.1-70B | All | 26.4 | 36.1 | 31.9 | 13.9 | 36.1 | 30.6 / 22.2 / 1.4 |
| | | Agree | 48.3 | 20.0 | 22.7 | 18.2 | 36.4 | 22.7 / 36.4 / 0.0 |
| | | Disagree | 25.0 | 16.7 | 38.1 | 14.3 | 31.4 | 28.6 / 14.3 / 4.8 |
| | | Discuss | 0.0 | 0.0 | 25.0 | 25.0 | 16.7 | 37.5 / 12.5 / 0.0 |
| | | Unrelated | 76.5 | 70.0 | 38.1 | 4.8 | 63.1 | 38.1 / 19.0 / 0.0 |
| **Few-shot** | Claude-3.5-Sonnet | All | 82.9 | 87.4 | 52.8 | 40.3 | **87.8** | 4.2 / 2.8 / 0.0 |
| | | Agree | 95.2 | 100.0 | 68.2 | 31.8 | 98.4 | 0.0 / 0.0 / 0.0 |
| | | Disagree | 97.6 | 100.0 | 28.6 | 66.7 | 97.6 | 4.8 / 0.0 / 0.0 |
| | | Discuss | 76.9 | 80.0 | 37.5 | 37.5 | 77.3 | 12.5 / 12.5 / 0.0 |
| | | Unrelated | 89.5 | 86.7 | 66.7 | 23.8 | 88.9 | 4.8 / 4.8 / 0.0 |
| | GPT-4o | All | 75.4 | 75.3 | 86.1 | 1.4 | 79.4 | 12.5 / 0.0 / 0.0 |
| | | Agree | 84.2 | 88.9 | 86.4 | 0.0 | 86.5 | 13.6 / 0.0 / 0.0 |
| | | Disagree | 97.6 | 97.6 | 90.5 | 4.8 | 96.8 | 4.8 / 0.0 / 0.0 |
| | | Discuss | 66.7 | 28.6 | 62.5 | 0.0 | 52.6 | 37.5 / 0.0 / 0.0 |
| | | Unrelated | 92.3 | 90.9 | 90.5 | 0.0 | 91.2 | 9.5 / 0.0 / 0.0 |
| | Command-R-32B | All | 63.8 | 61.4 | 48.6 | 33.3 | 69.0 | 16.7 / 1.4 / 0.0 |
| | | Agree | 84.2 | 86.5 | 27.3 | 54.5 | 84.2 | 18.2 / 0.0 / 0.0 |
| | | Disagree | 76.5 | 66.7 | 33.3 | 47.6 | 74.7 | 14.3 / 4.8 / 0.0 |
| | | Discuss | 40.0 | 0.0 | 62.5 | 12.5 | 38.3 | 25.0 / 0.0 / 0.0 |
| | | Unrelated | 97.6 | 94.7 | 81.0 | 4.8 | 92.7 | 14.3 / 0.0 / 0.0 |
| | Llama-3.1-70B | All | 35.2 | 44.9 | 19.4 | 12.5 | 37.3 | 47.2 / 20.8 / 0.0 |
| | | Agree | 42.9 | 54.5 | 31.8 | 13.6 | 47.6 | 36.4 / 18.2 / 0.0 |
| | | Disagree | 64.5 | 76.2 | 28.6 | 19.0 | 62.8 | 52.4 / 0.0 / 0.0 |
| | | Discuss | 22.2 | 0.0 | 0.0 | 12.5 | 11.6 | 50.0 / 37.5 / 0.0 |
| | | Unrelated | 55.2 | 40.0 | 4.8 | 4.8 | 34.9 | 52.4 / 38.1 / 0.0 |

Table 6: Human evaluation results (**part one**): **F1-2G** denotes the F1 score of the suggested class against the gold stance, **F1-2P** represents the F1 score of the suggested class against the predicted stance label of the model, **PEP** stands for the percentage of perfect explanations for each model, **IGCE** represents the percentage of incomplete generation but complete explanations, **OES** denotes Overall Explanation Score, and **CEP** indicates the completeness error percentage including the percentage of *Missed Details, Including Hallucination*, and *Empty Explanation*, respectively. **Human PE** stands for *Human performance estimate*. Refer to Section 5.2 for the definitions of these criteria. The worst and best results for each model per class, based on the OES, are highlighted in red and green, respectively.

| Model | Class | F1-2G | F1-2P | PEP | IGCE | OES | CEP |
|---|---|---|---|---|---|---|---|
| Aya-23-8B (PEFT) | All | 74.2 | 61.5 | 66.7 | 12.5 | **71.6** | 20.8 / 0.0 / 0.0 |
| | Agree | 90.0 | 92.3 | 63.6 | 22.7 | 89.5 | 13.6 / 0.0 / 0.0 |
| | Disagree | 80.0 | 78.8 | 61.9 | 9.5 | 76.7 | 28.6 / 0.0 / 0.0 |
| | Discuss | 66.7 | 0.0 | 25.0 | 12.5 | 34.7 | 62.5 / 0.0 / 0.0 |
| | Unrelated | 100.0 | 95.0 | 90.5 | 4.8 | 96.8 | 4.8 / 0.0 / 0.0 |
| *Similarity-based* | All | 62.2 | 53.0 | 30.6 | 2.8 | 49.5 | **62.5** / 4.2 / 0.0 |
| | Agree | 87.2 | 83.3 | 63.6 | 9.1 | 81.1 | 22.7 / 4.5 / 0.0 |
| | Disagree | 55.2 | 43.5 | 33.3 | 0.0 | 44.0 | 61.9 / 4.8 / 0.0 |
| | Discuss | 66.7 | 57.1 | 12.5 | 0.0 | 45.4 | 75.0 / 12.5 / 0.0 |
| | Unrelated | 95.0 | 97.4 | 0.0 | 0.0 | 64.1 | 100.0 / 0.0 / 0.0 |
| *Human PE.* | All | 71.9 | 66.0 | 69.4 | 0.0 | 69.1 | 30.6 / 0.0 / 0.0 |
| | Agree | 90.0 | 87.2 | 81.8 | 0.0 | 86.3 | 18.2 / 0.0 / 0.0 |
| | Disagree | 83.3 | 82.4 | 66.7 | 0.0 | 77.5 | 33.3 / 0.0 / 0.0 |
| | Discuss | 66.7 | 33.3 | 25.0 | 0.0 | 41.7 | 75.0 / 0.0 / 0.0 |
| | Unrelated | 92.3 | 94.1 | 76.2 | 0.0 | 87.5 | 23.8 / 0.0 / 0.0 |
| *Human* | All | 87.6 | 87.6 | 90.3 | 0.0 | **88.5** | 8.3 / 1.4 / 0.0 |
| | Agree | 100.0 | 100.0 | 100.0 | 0.0 | 100.0 | 0.0 / 0.0 / 0.0 |
| | Disagree | 97.6 | 97.6 | 90.5 | 0.0 | 95.2 | 4.8 / 4.8 / 0.0 |
| | Discuss | 66.7 | 66.7 | 37.5 | 0.0 | 57.0 | 62.5 / 0.0 / 0.0 |
| | Unrelated | 100.0 | 100.0 | 100.0 | 0.0 | 100.0 | 0.0 / 0.0 / 0.0 |

Table 7: Human evaluation results (**part two**): **F1-2G** denotes the F1 score of the suggested class against the gold stance, **F1-2P** represents the F1 score of the suggested class against the predicted stance label of the model, **PEP** stands for the percentage of perfect explanations for each model, **IGCE** represents the percentage of incomplete generation but complete explanations, **OES** denotes Overall Explanation Score, and **CEP** indicates the completeness error percentage including the percentage of *Missed d=Details, Including Hallucination*, and *Empty Explanation*, respectively. **Human PE** stands for *Human performance estimate*. Refer to Section 5.2 for the definitions of these criteria. The worst and best results for each model per class, based on the OES, are highlighted in red and green, respectively.

| | |
|---|---|
| **Example** | |
| **Claim:** | |
| عدم دعوت جو بایدن، رئیس جمهوری ایالات متحده از ورزشکاران آمریکایی در بازی‌های المپیک توکیو به کاخ سفید. | |
| American athletes from the Tokyo Olympics have not been invited by U.S. President Joe Biden to visit the White House. | |
| **Context:** | |
| جو بایدن، رئیس جمهوری ایالات متحده و همسرش جیل بایدن بانوی نخست کشور، ورزشکاران آمریکایی حاضر در بازی‌های المپیک توکیو را به دلیل نشان دادن شجاعت در شرایط سخت، و متحد کردن کشور مورد ستایش قرار دادند. جو بایدن، رئیس جمهوری ایالات متحده و همسرش جیل بایدن بانوی نخست کشور، ورزشکاران آمریکایی حاضر در بازی‌های المپیک توکیو را به دلیل نشان دادن شجاعت در شرایط سخت، و متحد کردن کشور مورد ستایش قرار دادند. به گزارش رویترز، آنها پیشتر نیز دعوت این ورزشکاران به کاخ سفید را نیز داده بودند. ایالات متحده با ۶۱۳ ورزشکار در بازی‌های توکیو شرکت کرد که دومین کاروان بزرگ ورزشی آمریکا در تاریخ المپیک محسوب می‌شود. این کشور در مجموع با کسب ۱۱۳ مدال، از تمامی رقبای خود پیشی گرفت و صدرنشین جدول مدال‌ها شد. جو بایدن در تماس تصویری با ورزشکاران و خانواده‌های آنها گفت «این فقط توانایی ورزشی شما نبود، بلکه شجاعت اخلاقی شما بود. شما به ما یادآوری کردید که چه کشور شگفت‌انگیزی هستیم، و باعث شدید تا به عنوان یک کشور، بسیار خوب به نظر برسیم». آقای بایدن همچنین از چندین ورزشکار به صورت مشخص قدردانی کرد که از آنها سیمون بایلز، ژیمناست آمریکایی که به دلیل نگرانی از سلامت روان، از تعدادی از رقابت‌های انفرادی و تیمی کناره‌گیری کرد، ولی در نهایت موفق به کسب یک مدال برنز شد. رئیس جمهور ایالات متحده از ادامه آیزایا جوئت، دونده آمریکایی را نیز ستود که در یک رقابت دو در موقعیت خوبی بود، اما در برخورد غیرعمدی با نایجل آموس از بوتسوانا، هر دو ورزشکار زمین خوردند. او در اقدامی تحسین‌برانگیز، به دونده بوتسوانا کمک کرد تا بلند شود و همراه با یکدیگر از خط پایان گذشتند. جیل بایدن، بانوی نخست آمریکا که به نمایندگی از ایالات متحده و به منظور حمایت از ورزشکاران این کشور به توکیو سفر کرده بود، گفت که «آمریکا به شفا نیاز دارد و این رویداد، شادی زیادی را به همراه داشت.» او در ادامه افزود که «ورزشکاران حاضر در این بازی‌ها، آمریکایی‌ها را گردهم آوردند و ما را شاد کردند، از این بابت از آنها سپاسگزارم.» | |
| U.S. President Joe Biden and First Lady Jill Biden praised American athletes who participated in the Tokyo Olympics for courage in challenging circumstances and uniting the country. U.S. President Joe Biden and First Lady Jill Biden praised American athletes who participated in the Tokyo Olympics for courage in challenging circumstances and uniting the country. The United States competed in the Tokyo Games with 613 athletes, forming the second-largest U.S. Olympic delegation in history. The country secured a total of 113 medals, surpassing all competitors to top the medal table. In a video call with the athletes and their families, Joe Biden said, "It wasn't just your athletic abilities; it was your moral courage. You reminded us of what an amazing country we are and made us look good as a nation." | |
| President Biden also specifically acknowledged several athletes, including American gymnast Simone Biles, who withdrew from some individual and team events due to mental health concerns but ultimately won a bronze medal. He also praised American runner Isaiah Jewett, who was in a good position in a race but fell after an accidental collision with Nigel Amos of Botswana. In an admirable gesture, Jewett helped Amos get up, and they crossed the finish line together. | |
| First Lady Jill Biden, who traveled to Tokyo on behalf of the United States to support the athletes, stated, "America needs healing, and this event brought a lot of joy." She added, "The athletes in these games brought Americans together and made us happy, and for that, I am grateful to them." | |
| **Gold class:** Disagree | |
| **Claude-3.5-Sonnet (Zero)** | |
| **Explanation:** | |
| توضیح به فارسی: بر اساس متن ارائه شده، این ادعا نادرست است. در متن به صراحت ذکر شده است که جو بایدن و همسرش جیل بایدن، ورزشکاران آمریکایی حاضر در بازی‌های المپیک توکیو را به کاخ سفید دعوت کرده‌اند. متن می‌گوید: "به گزارش رویترز، آن | |
| Explanation in Farsi: Based on the provided text, this claim is incorrect. The text clearly states that Joe Biden and his wife, Jill Biden, invited the American athletes who participated in the Tokyo Olympics to the White House. The text mentions, "According to Reuters, they ... " | |
| **Predicted class:** Disagree | |
| **GPT-4o (Zero)** | |
| **Explanation:** | |
| جو بایدن، رئیس جمهوری ایالات متحده و همسرش جیل بایدن بانوی نخست کشور، ورزشکاران آمریکایی حاضر در بازی های المپیک توکیو را به کاخ سفید دعوت کرده بودند. | |
| U.S. President Joe Biden and First Lady Jill Biden had invited the American athletes who participated in the Tokyo Olympics to the White House. | |
| **Predicted class:** Disagree | |
| **Command-R-32B (Zero)** | |
| **Explanation:** | |
| جو بایدن و همسرش ورزشکاران آمریکایی را به کاخ سفید دعوت کردند. | |
| Joe Biden and his wife invited the American athletes to the White House. | |
| **Predicted class:** Disagree | |
| **Llama-3.1-70B (Zero)** | |
| **Explanation:** | |
| جو بایدن، رئیس جمهوری ایالات متحده و همسرش جیل بایدن بانوی نخست کشور، ورزشکاران آمریکایی حاضر در بازی های المپیک توکیو را به دلیل نشان دادن شجاعت در شرایط سخت، و متحد کردن کشور مورد ستایش قرار دادند. به گزارش رویترز، آنها پیشتر این ورزشکاران را نیز به کاخ سفید دعوت کرده بودند. ایالات متحده با ۶۱۳ ورزشکار در بازی های توکیو شرکت کرد که دومین کاروان بزرگ ورزشی آمریکا در تاریخ المپیک محسوب می شود. این کشور در مجموع با کسب ۱۱۳ مدال، از تمامی رقبای خود پیشی گرفت و صدرنشین جدول مدال ها شد. جو | |
| U.S. President Joe Biden and First Lady Jill Biden praised the American athletes who participated in the Tokyo Olympics for demonstrating courage under challenging circumstances and for uniting the nation. According to Reuters, they had previously invited these athletes to the White House as well. The United States competed in the Tokyo Games with 613 athletes, marking the second-largest U.S. Olympic delegation in history. The country won a total of 113 medals, surpassing all competitors to lead the medal table. Joe | |
| **Predicted class:** Disagree | |
| **Claude-3.5-Sonnet (Few)** | |
| **Explanation:** | |
| توضیح به فارسی: بر اساس متن، جو بایدن و همسرش جیل بایدن، ورزشکاران آمریکایی حاضر در بازی‌های المپیک توکیو را به کاخ سفید دعوت کرده‌اند. در متن به صراحت آمده است: "آنها پیشتر این ورزشکاران را به کاخ سفید نیز دعوت کرده بودند." بنابراین، ادعای عدم دعوت جو بایدن از ورزشکاران آمریکایی به کاخ سفید با اطلاعات موجود در متن در تضاد است و نادرست می‌باشد. | |
| Explanation in Farsi: Based on the text, Joe Biden and his wife, Jill Biden, invited the American athletes who participated in the Tokyo Olympics to the White House. The text clearly states: "They had previously invited these athletes to the White House." Therefore, the claim that Joe Biden did not invite the American athletes to the White House contradicts the information in the text and is incorrect. | |
| **Predicted class:** Disagree | |
| **GPT-4o (Few)** | |
| **Explanation:** | |
| جو بایدن، رئیس جمهوری ایالات متحده و همسرش جیل بایدن بانوی نخست کشور، ورزشکاران آمریکایی حاضر در بازی های المپیک توکیو را به دلیل نشان دادن شجاعت در شرایط سخت، و متحد کردن کشور مورد ستایش قرار دادند. به گزارش رویترز، آنها پیشتر این ورزشکاران را به کاخ سفید نیز دعوت کرده بودند. | |
| U.S. President Joe Biden and First Lady Jill Biden praised the American athletes who participated in the Tokyo Olympics for demonstrating courage in challenging circumstances and for uniting the country. According to Reuters, they had previously invited these athletes to the White House as well. | |
| **Predicted class:** Disagree | |
| **Command-R-32B (Few)** | |
| **Explanation:** | |
| جو بایدن ، رئیس جمهوری ایالات متحده و همسرش جیل بایدن بانوی نخست کشور، ورزشکاران آمریکایی حاضر در بازی های المپیک توکیو را به دلیل نشان دادن شجاعت در شرایط سخت ، و متحد کردن کشور مورد ستایش قرار دادند . به گزارش رویترز، آنها پیشتر این ورزشکاران را به کاخ سفید نیز دعوت کرده بودند . ایالات متحده با ۶۱۳ ورزشکار در بازی های توکیو شرکت کرد که دومین کاروان بزرگ ورزشی آمریکا در تاریخ المپیک محسوب می شود . این کشور در مجموع با کسب ۱۱۳ مدال ، از تمامی رقبای خود پیشی گرفت و صدرنشین جدول مدال ها شد . جو بایدن در تماس تصو | |
| U.S. President Joe Biden and First Lady Jill Biden praised the American athletes who participated in the Tokyo Olympics for showing courage in difficult circumstances and for uniting the country. According to Reuters, they had previously invited these athletes to the White House as well. The United States competed in the Tokyo Games with 613 athletes, marking the second-largest U.S. Olympic delegation in history. The country won a total of 113 medals, surpassing all competitors to top the medal table. Joe Biden, in a video call with the athletes and their families, said... | |
| **Predicted class:** Unrelated | |
| **Llama-3.1-70B (Few)** | |
| **Explanation:** | |
| جو بایدن، رئیس جمهوری ایالات متحده و همسرش جیل بایدن بانوی نخست کشور، ورزشکاران آمریکایی حاضر در بازی های المپیک توکیو را به دلیل نشان دادن شجاعت در شرایط سخت، و متحد کردن کشور مورد ستایش قرار دادند. به گزارش رویترز، آنها پیشتر این ورزشکاران را به کاخ سفید نیز دعوت کرده بودند. ایالات متحده با ۶۱۳ ورزشکار در بازی های توکیو شرکت کرد که دومین کاروان بزرگ ورزشی آمریکا در تاریخ المپیک محسوب می شود. این کشور در مجموع با کسب ۱۱۳ مدال، از تمامی رقبای خود پیشی گرفت و صدرنشین جدول مدال ها شد. جو | |
| U.S. President Joe Biden and First Lady Jill Biden praised the American athletes who participated in the Tokyo Olympics for their courage in difficult circumstances and for uniting the country. According to Reuters, they had previously invited these athletes to the White House as well. The United States participated in the Tokyo Games with 613 athletes, marking the second-largest U.S. Olympic delegation in history. The country won a total of 113 medals, surpassing all competitors to lead the medal table. Joe | |
| **Predicted class:** Agree | |
| **Aya-23-8B (PEFT)** | |
| **Explanation:** | |
| جو بایدن، رئیس جمهوری ایالات متحده و همسرش جیل بایدن بانوی نخست کشور، ورزشکاران آمریکایی حاضر در بازی های المپیک توکیو را به دلیل نشان دادن شجاعت در شرایط سخت ، و متحد کردن کشور مورد ستایش قرار دادند . جو بایدن ، رئیس جمهوری ایالات متحده و همسرش جیل بایدن بانوی نخست کشور، ورزشکاران آمریکایی حاضر در بازی های المپیک توکیو را به دلیل نشان دادن شجاعت در شرایط سخت ، و متحد کردن کشور مورد ستایش قرار دادند . به گزارش رویترز، آنها پیشتر این ورزشکاران را به کاخ سفید نیز دعوت کرده‌بودند. | |
| U.S. President Joe Biden and First Lady Jill Biden praised the American athletes who participated in the Tokyo Olympics for their courage in difficult circumstances and for uniting the country. U.S. President Joe Biden and First Lady Jill Biden praised the American athletes who participated in the Tokyo Olympics for their courage in difficult circumstances and for uniting the country. According to Reuters, they had previously invited these athletes to the White House as well. | |
| **Predicted class:** Disagree | |
| **Similarity-based** | |
| **Explanation:** | |
| جو بایدن، رئیس جمهوری ایالات متحده و همسرش جیل بایدن بانوی نخست کشور، ورزشکاران آمریکایی حاضر در بازی‌های المپیک توکیو را به دلیل نشان دادن شجاعت در شرایط سخت، و متحد کردن کشور مورد ستایش قرار دادند. جو بایدن، رئیس جمهوری ایالات متحده و همسرش جیل بایدن بانوی نخست کشور، ورزشکاران آمریکایی حاضر در بازی‌های المپیک توکیو را به دلیل نشان دادن شجاعت در شرایط سخت، و متحد کردن کشور مورد ستایش قرار دادند. | |
| U.S. President Joe Biden and First Lady Jill Biden praised the American athletes who participated in the Tokyo Olympics for their courage in difficult circumstances and for uniting the country. U.S. President Joe Biden and First Lady Jill Biden praised the American athletes who participated in the Tokyo Olympics for their courage in difficult circumstances and for uniting the country. | |
| **Predicted class:** Disagree | |
| **Human PE** | |
| **Explanation:** | |
| به گزارش رویترز، آن ها پیشتر این ورزشکاران را به کاخ سفید نیز دعوت کرده بودند. | |
| According to Reuters, they had previously invited these athletes to the White House as well. | |
| **Predicted class:** Disagree | |

Figure 8: The generated explanations of different LLMs. The Human-provided gold explanation is highlighted in green.